\documentclass[11pt]{article}

\usepackage[final]{acl}

\usepackage{times}
\usepackage{latexsym}
\usepackage{enumitem}
\usepackage{amsmath}
\usepackage{setspace}
\usepackage{multirow}

\usepackage[T1]{fontenc}

\usepackage[utf8]{inputenc}

\usepackage{microtype}

\usepackage{inconsolata}

\usepackage{graphicx}

\title{Efficient Prompt Optimisation for Legal Text Classification with Proxy Prompt Evaluator}

\author{Hyunji Lee$^{1*}$,   Kevin Li$^{1*}$,  Matthias Grabmair$^1$, Shanshan Xu$^{1,2,3}$ \\
$^{1}$Technical University of Munich, Germany \\
$^{2}$Department of Computer Science, University of Copenhagen, Copenhagen, Denmark\\
$^{3}$Faculty of Law, University of Copenhagen, Copenhagen, Denmark\\
 \texttt{\{hyunji39.lee, kevinchenhao.li, matthias.grabmair\}@tum.de} \\
  \texttt{shanshan.xu@di.ku.dk}}

\begin{document}
\maketitle
\begin{abstract}
Prompt optimization aims to systematically refine prompts to enhance a language model’s performance on specific tasks. Fairness detection in Terms of Service (ToS) clauses is a challenging legal NLP task that demands carefully crafted prompts to ensure reliable results. However, existing prompt optimization methods are often computationally expensive due to inefficient search strategies and costly prompt candidate scoring. In this paper, we propose a framework that combines Monte Carlo Tree Search (MCTS) with a proxy prompt evaluator to more effectively explore the prompt space while reducing evaluation costs. Experiments demonstrate that our approach achieves higher classification accuracy and efficiency than baseline methods under a constrained computation budget.
\end{abstract}

\def\thefootnote{*}\footnotetext{These authors contributed equally to this work}
\section{Introduction}

Terms of Service (ToS) agreements are lengthy, complex documents that define the legal relationship between companies and consumers. While these documents are critical for protecting consumer rights and regulating corporate practices, clauses in ToS agreements are often written in highly complex language, making them difficult for the users to understand. As a result, unfair or potentially exploitative ToS clauses, may go unnoticed. Detecting such unfair clauses is therefore essential for promoting transparency, consumer protection, and regulatory compliance. 

Manual review of such documents is however extremely time-consuming and requires legal knowledge. Large language models (LLMs) therefore offer a promising alternative by automatically classifying unfair ToS clauses at scale. Nevertheless, the performance of LLMs is highly sensitive to the prompt design. Prior work has shown that even minor variation in prompt wording and formatting can substantially affect accuracy and consistency \cite{salinas2024butterflyeffectalteringprompts, he2024doespromptformattingimpact}.

Recently, there is growing research interest in prompt optimization, which is the process of systematically refining prompts to improve a language model’s performance on a specific task \cite{prasad2023gripsgradientfreeeditbasedinstruction, pryzant2023automaticpromptoptimizationgradient, yang2024largelanguagemodelsoptimizers, ma2024largelanguagemodelsgood, choi2025promptoptimizationmetalearning, xiang2025selfsupervisedpromptoptimization}. Prompt optimization is typically framed as an iterative search process that involves modules such as generating revised candidate prompts, evaluating prompt performance, and searching for the best-performing candidates to guide subsequent refinements.

Despite recent advances, most optimization methods still struggle with inefficient exploration of candidate prompts space. For example, beam search, a widely used search strategy in prior work \cite{pryzant2023automaticpromptoptimizationgradient}, often produces repetitive and untargeted edits, relying on costly deterministic forward-only search. In this work, we adopt Monte Carlo Tree Search (MCTS) \cite{coulom2006efficient}, inspired by PromptAgent \cite{wang2023promptagentstrategicplanninglanguage} to improve the exploration efficiency of the candidate prompts space. MCTS strategically models the search space as a tree and updates future reward estimates through backpropagation.


Another major bottleneck in current prompt optimization methods is the high computational cost of evaluating candidate prompts. Each evaluation typically requires costly inference on LLMs and is repeated across a large pool of candidates. To reduce computation, most methods assess prompt performance using a small score set, a subset sampled from the full validation set. While this approach is faster and cheaper, the small size of the subset can cause performance estimates to fluctuate depending on which samples are included. Moreover, prompts optimized on a small score set may fail to generalize well to the full dataset.

To mitigate this, we augment our framework with a proxy prompt evaluator based on a correctness classifier, inspired by PromptEval \cite{polo2024efficientmultipromptevaluationllms}. This proxy scorer efficiently evaluates the performance of prompt candidates by predicting their correctness on the target task, reducing the need for repeatedly calling costly LLM and therefore enabling  evaluation of the prompts across the full validation set.

Our prompt optimization framework enables efficient exploration of the prompt search space and streamlines evaluation. Our results show that the MCTS approach discovers better-performing prompts than existing optimization frameworks and, when combined with a proxy prompt evaluator, achieves similar binary classification performance with reduced computational cost.

\section{Related Work}
\label{sec:relwork}

\subsection{Unfair ToS Clause Detection}

The detection of unfair clauses in ToS documents has been an active line of research in legal natural language processing. A prominent benchmark in this area is the CLAUDETTE dataset, which contains annotated clauses from consumer contracts labeled as \textit{fair} or \textit{unfair}. \citet{Lippi_2019} first introduced this dataset and developed methods for unfairness detection using machine learning techniques to support consumer protection. Subsequent work by \citet{ruggeri2022detecting} extended this line of research, refining both the dataset and the detection methods to improve robustness and applicability using memory-augmented neural networks. Nevertheless, later work on adversarial attacks have exposed a significant weakness: these classification systems are highly sensitive to perturbations in input phrasing \cite{xu2022attack},  questioning their practical reliability. These findings highlighted the sensitivity of legal text classification models and motivated further research into methods for improving robustness.

\subsection{Prompt Optimization}

The general process of prompt optimization can be split into the following key modules: prompt update, search strategy and prompt evaluation.

\subsubsection{Prompt Update}
The prompt updating methods used in prior work primarily fall into three categories: resampling-based, explicit reflection-based, and implicit reflection-based \cite{ma2024largelanguagemodelsgood}.

\paragraph{Resampling-based} approaches apply random edit operations (e.g. deletion, swap, paraphrase, addition) to the base prompt without directional feedback. For example, GrIPS \cite{prasad2023gripsgradientfreeeditbasedinstruction} repeatedly generates candidate prompts via such edits, evaluates them on a held-out set, and selects the best-performing one. However, the lack of guidance often leads to ineffective edits and poor performance.

\paragraph{Implicit reflection-based} approaches, such as OPRO \cite{yang2024largelanguagemodelsoptimizers}, generate new prompts based on the history of candidate prompts and their performance scores. However, these methods do not require the prompt optimizer to explicitly reflect on the errors of previous prompts. While this approach is more guided than simple resampling strategies, it still lacks direct feedback mechanisms that consider the nature of past mistakes.

\paragraph{Explicit reflection-based} approaches incorporate natural language feedback as \textit{textual gradients} to guide edits. ProTeGi \cite{pryzant2023automaticpromptoptimizationgradient} exemplifies this idea by using an LLM to identify weaknesses in a prompt and propose semantic edits in the opposite direction. While more effective, recent work indicates that such methods produce repetitive feedback and often struggle to align improvements in prompt text with downstream model behavior \cite{ma2024largelanguagemodelsgood}.

\subsubsection{Search Strategy}
The search strategy decides which prompt candidates are selected, filtered and further expanded. Common strategies include the following:

\paragraph{Greedy search} is the simplest approach, where only the highest-scoring prompt from the current iteration is selected for expansion in the next step, for example used by OPRO \cite{yang2024largelanguagemodelsoptimizers}. While computationally efficient, it risks premature convergence because potentially better prompts in the search space are not reached.

\paragraph{Beam search} maintains a beam, consisting of top-performing prompts at each iteration, expanding all of them in parallel, such as ProTeGi \cite{pryzant2023automaticpromptoptimizationgradient} and GrIPS \cite{prasad2023gripsgradientfreeeditbasedinstruction}. This allows it to explore multiple promising paths simultaneously, reducing the chance of missing promising prompts. However, the beam width is an important parameter, as a narrow beam can still miss high-performing prompts, while a wide beam increases computational cost.

\begin{figure*}[t!]
  \includegraphics[width=455pt]{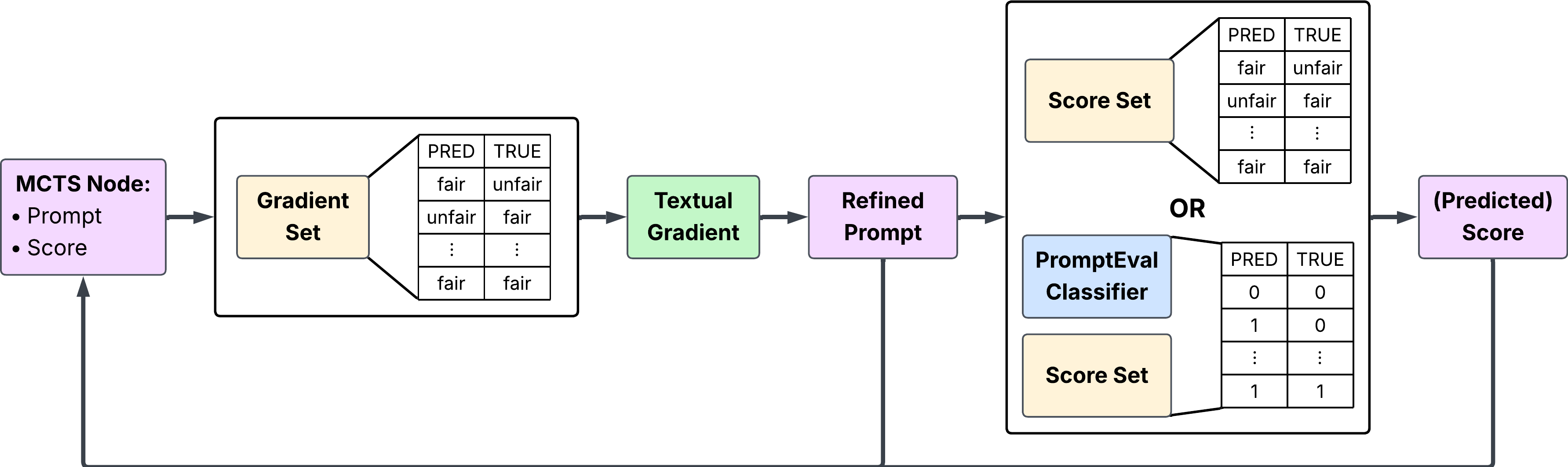} 
  \caption {Our framework with scoring on the score set or alternative scoring with the trained PromptEval-based correctness classifier.}
  \label{fig:mcts_fw}
\end{figure*}

\paragraph{Tree-based search} strategies, such as Monte Carlo Tree Search (MCTS) \cite{wang2023promptagentstrategicplanninglanguage}, explicitly represent the space of prompt candidates as a tree. The prior approaches often use a deterministic forward-only search strategy for choosing the next prompt candidate, which limits their ability to revisit and select the most promising prompts in the search space. In contrast, MCTS is a search algorithm that balances exploration and exploitation and revisits earlier prompts. This allows MCTS to identify better prompts on alternative tree paths and potentially outperform the current prompt candidate.

\subsubsection{Prompt Evaluation}

A major source of cost in prompt optimization arises from repeatedly querying an LLM on a evaluation set at every optimization step to assess prompt performance. The challenge of high computation cost due to repeatedly calling the LLM is not unique to prompt optimization. Recently an increasing amount of research has been done on predicting performance without running the full inference \cite{beyer2025fastproxiesllmrobustness, berrada2025scalingactivetestinglarge, zhong2025efficientevaluationlargelanguage}.

PromptEval \cite{polo2024efficientmultipromptevaluationllms} addresses this issue by introducing a lightweight model to predict the performance of a given prompt on a specific task. In this work, we train a prompt performance prediction model and use it as a proxy prompt evaluation module, which enables fast and efficient prompt performance evaluations without requiring costly LLM inference on the whole evaluation set.




\section{Dataset}

\begin{figure}[h]
\centering
\includegraphics[width=\columnwidth]{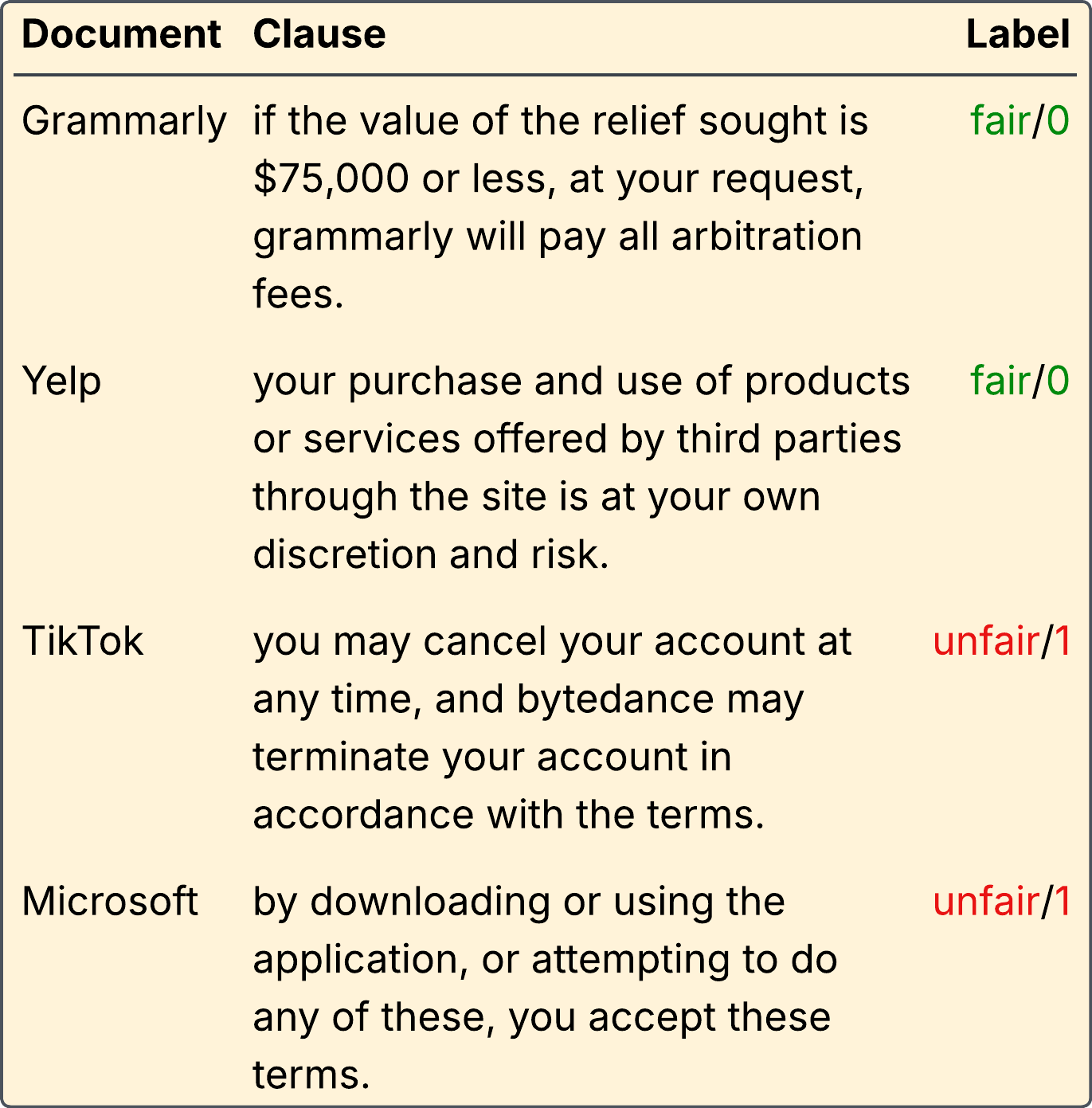}
\caption{Examples from the CLAUDETTE dataset.}
\label{fig:CLAUDETTE_ex}
\end{figure}

We conduct our prompt optimization experiments on the CLAUDETTE dataset \cite{Lippi_2019}, which contains 20,417 clauses extracted from 50 ToS contracts of global online platforms. Each clause was manually labeled by legal experts as \textit{fair} or \textit{unfair}.\footnote{In the original CLAUDETTE dataset from \citealt{Lippi_2019}, the ToS clauses are annotated in three labels: 1 standing for \textit{clearly fair}, 2 for \textit{potentially unfair}, and 3 for \textit{clearly unfair}. In our work, we merged the label of \textit{potentially unfair} and \textit{clearly unfair} to \textit{unfair}.} A clause is labeled \textit{unfair} if it somehow introduces an unacceptable imbalance in the parties’ rights and obligations, i.e., harms the user’s rights or minimizes the online service’s obligations. In addition, each unfair clause is annotated with one or more of nine unfairness categories (e.g., arbitration, content removal, jurisdiction) depending on the source of the unfairness. Figure \ref{fig:CLAUDETTE_ex} exhibits example ToS clauses for CLAUDETTE. Notably, the label ratio in  CLAUDETTE is extremely imbalanced, with a  distribution of roughly 9:1 (\textit{fair}:\textit{unfair}), as displayed in Table \ref{table:CLAUDETTE_stats}.



\begin{table}[h]
\centering
\begin{tabular}{lccc} 
 \hline
 Split & \# of clauses & \% fair & \% unfair\\
 \hline
 Train & 8,354 & 89.5\% & 10.5\%\\
 Val & 8,279 & 89.1\% & 10.9\%\\ 
 Test & 3,784 & 89.3\% & 10.7\%\\
 \hline
\end{tabular}%
\caption{CLAUDETTE statistics.}
\label{table:CLAUDETTE_stats}
\end{table}

\section{Methodology}
\label{sec:methodology}
We investigate the performance of our prompt optimization framework for the task of unfair ToS clause detection. We begin with a simple initial prompt \textit{"Is this clause fair (0) or unfair (1) to the consumer?"}\label{init_prompt}. The key modules of our prompt optimization process is illustrated in Figure \ref{fig:mcts_fw}. Specifically, we update the prompt and generate various prompt candidates using \textit{textual gradients} \cite{pryzant2023automaticpromptoptimizationgradient} (\S \ref{sec:prompt_update}). To efficiently search among candidate prompts, we employ Monte Carlo Tree Search (MCTS) following \citealt{wang2023promptagentstrategicplanninglanguage} (\S \ref{sec:mcts}). A major bottleneck of this approach is the high computational cost when evaluating among candidate prompts. To mitigate this, we propose training an external prompt grader model (PromptEval \cite{polo2024efficientmultipromptevaluationllms}) as a proxy selection module (\S \ref{sec:prompteval}).

\begin{figure}[h]
\centering
\includegraphics[width=0.99\columnwidth]{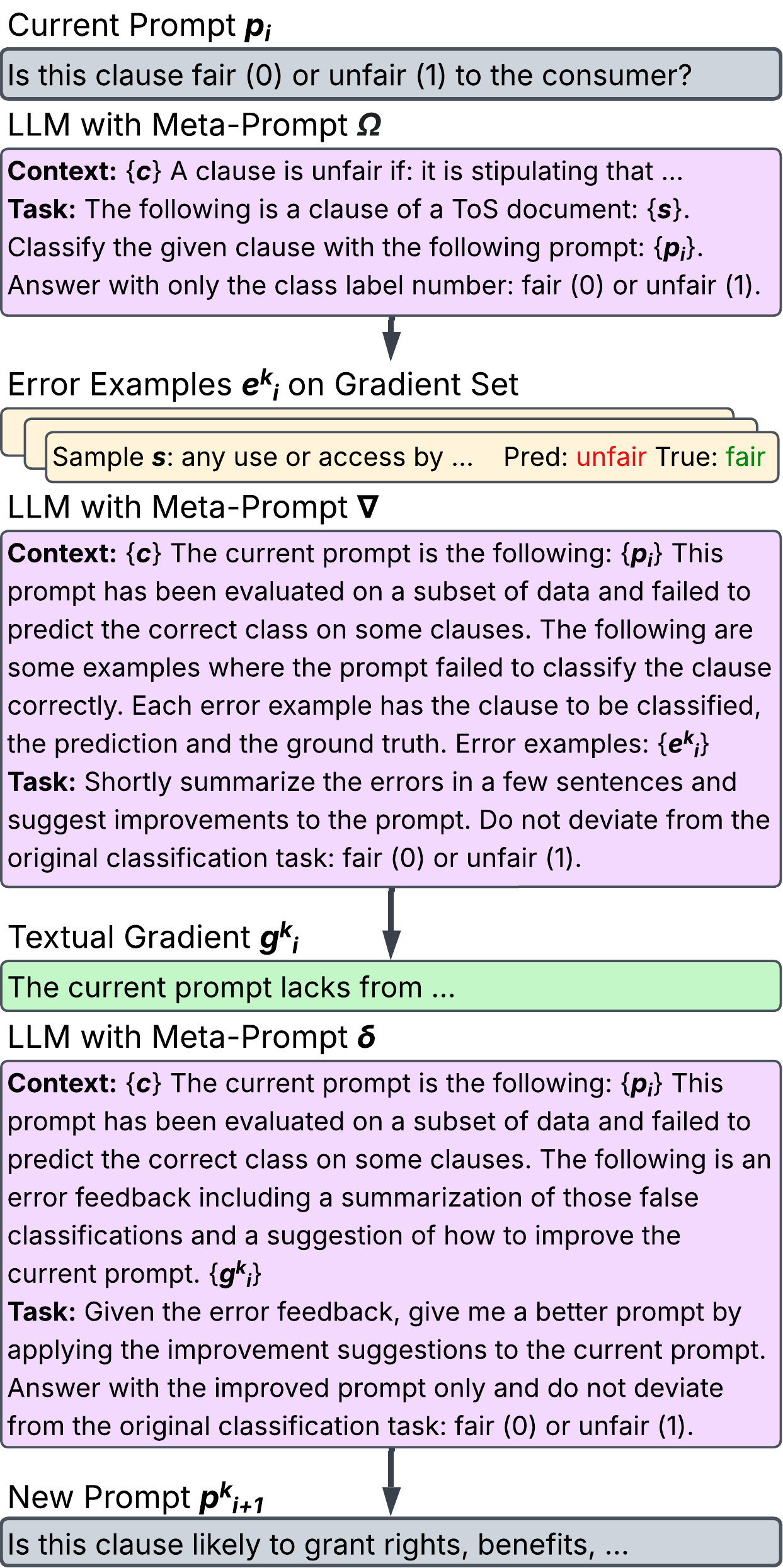}
\caption{Prompt update with textual gradients. The global context \textit{\textbf{c}} is \textit{"You are a prompt optimizer for legal documents. The task is to classify clauses of Terms of Service documents according to the given prompt."}.}
\label{fig:text_grad}
\end{figure}

\subsection{Prompt Update with Textual Gradient}
\label{sec:prompt_update}

We updated the prompts using \textit{textual gradients} \cite{pryzant2023automaticpromptoptimizationgradient}. The term "textual gradient" refers to a natural language-based feedback mechanism that require an LLM as prompt optimizer, to critique the current prompt's performance and suggests improvements. This feedback meachnism is analogous to how numerical gradients guide optimization in machine learning. 

At each iteration $i$, we queried Llama-3-8B-Instruct \cite{llama3modelcard} through DeepInfra's API (version dated 2024-04-18) to evaluate the current prompt $p_i$ on a minibatch $m^k_i$ of dataset samples (hereafter referred to as the \textit{gradient set}) using the meta-prompt $\Omega$ (shown in Figure \ref{fig:text_grad}). This meta-prompt included a description of different types of unfair clauses mentioned in \citet{Lippi_2019}. To ensure structured outputs, it explicitly requested numerical responses, making the results easier to parse. The false predictions on the gradient set (hereafter referred to as error examples $e^k_i$) were then passed to another meta-prompt $\nabla$, which produced a natural language summary of the weaknesses in $p_i$ along with improvement suggestions. This feedback served as the \textit{textual gradient} $g^k_i$.  

We then applied the meta-prompt $\delta$ (detailed in Figure \ref{fig:text_grad}), combining the current prompt $p_i$ and the textual gradient $g^k_i$ to instruct the model to perform semantic edits that address the identified flaws. This process yielded a set of improved prompt candidates $p_{i+1}^k$, where $k$ denotes the number of candidates generated at each iteration (we used 4 candidates per iteration).  

For the gradient set, we randomly sampled 20 clauses from the training set, resampling at each iteration to ensure diverse feedback. The label distribution of \textit{fair} and \textit{unfair} was maintained at 55:45, with the unfair subset including 5\% from each of the nine multi-label unfairness categories.

\subsection{Prompt Search with MCTS}
\label{sec:mcts}

We followed the implementation of MCTS described in \citet{wang2023promptagentstrategicplanninglanguage}. MCTS is a search algorithm that explores candidate prompts by building a search tree. Each node in the tree represents a prompt with values such as visit counts and estimated performance. The process consists of four steps: selection (choosing a promising node), expansion (adding new nodes), simulation (running rollouts to estimate outcomes), and backpropagation (updating node values). By repeating this loop, MCTS balances exploring new possibilities with exploiting known effective prompts.  

For expansion and simulation, we used the prompt update method with textual gradients and the same meta-prompts to generate new prompts (\S \ref{sec:prompt_update}). Future performance of a prompt was estimated with a Q-value, similar to a Markov Decision Process. For Q-value estimation, we evaluated and scored each node’s prompt on a separate fixed batch of 200 random training samples (hereafter the \textit{score set}), which was drawn to match the distribution of the gradient set. The LLM was queried with the same meta-prompt $\Omega$ for scoring (see Figure \ref{fig:text_grad}).  

MCTS was run for 12 iterations, with 4 prompt candidate generations per iteration and a depth limit of 8, starting from the initial prompt as the root node at depth 0. An early stopping criterion with a patience of 5 was applied after each backpropagation step. For the performance evaluation on the score set, three different scoring metrics were used: macro F1, accuracy, and random scoring. We conducted five independent runs for each metric. For the model settings, we set the temperature to 0.0 when evaluating on the score set to reduce noise and improve consistency. We set the temperature to 1.0 during the generation of new prompts to increase prompt diversity.

\subsection{Proxy Prompt Evaluator}
\label{sec:prompteval}

To reduce computational cost in evaluating prompt performance for unfair ToS clause detection, we train a lightweight model based on PromptEval \cite{polo2024efficientmultipromptevaluationllms} to serve as a proxy evaluation module. This proxy acts as a fast estimator that predicts whether the LLM would classify each clause correctly under a given prompt.

Formally, given a ToS clause $x_j$, a prompt $p_i$, and the gold fairness label $y_j \in \{0,1\}$, 
the LLM ToS fairness classifier $f$ produces a prediction:
\[
\hat{y}_{i,j} = f(p_i, x_j).
\]

The correctness of this prediction is defined as:
\[
c_{i,j} = \mathbf{1}\{\hat{y}_{i,j} = y_j\}.
\]

To approximate this correctness signal efficiently, we train a proxy prompt evaluator $\phi$ as correctness classifier. 
Each training instance is represented as:
\[
z_{i,j} = \big[e(p_i) \;\Vert\; e(x_j) \;\Vert\;(y_j)\big],
\]

where $e(\cdot)$ is an embedding function and $\Vert$ denotes concatenation. The proxy prompt evaluator $\phi$ produces:

\[
\hat{c}_{i,j} = \phi(z_{i,j}) \in [0,1],
\]

which estimates the probability that the LLM classifier $f$ correctly predicts the fairness of clause $x_j$ under prompt $p_i$.

The proxy prompt evaluator $\phi$ is trained using binary cross-entropy loss, where $\theta$ are the parameters of $\phi$:

\[
\mathcal{L}(\theta) = - \sum_{i,j} \Big[ c_{i,j} \log \hat{c}_{i,j} + (1 - c_{i,j}) \log (1 - \hat{c}_{i,j}) \Big]. 
\]


By using the proxy promt evaluator $\phi$, we can evaluate candidate prompts over the entire validation set without repeated expensive calls to the LLM. In our experiments, the \textit{score set} increased from 200 (as used in the original MCTS method), to 8,279 samples, which is the full validation set of CLAUDETTE. Thereby we managed to improve search stability and reduce evaluation costs while keeping the overall optimization procedure unchanged. Moreover, the bigger size of the \textit{score set} may also lead to the better generalization of the improved prompts. To further improve efficiency, the system implements embedding caching: once a prompt, sample, or label embedding is computed, it is stored in memory and reused in future evaluations. Since many prompts are evaluated repeatedly across the search tree, this avoids redundant computations and significantly reduces total runtime.

\subsubsection{Constructing the Correctness Dataset}

To train the proxy model, we required a dataset that records when the LLM binary classifier succeeds or fails at fairness prediction under different prompts. This correctness dataset is built by pairing candidate prompts with clauses from the CLAUDETTE dataset, comparing the LLM’s deterministic predictions to the gold labels, and assigning a binary correctness indicator.

Each entry consists of: (1) an embedding of the prompt, (2) an embedding of the clause, (3) a one-hot encoding of the gold fairness label, and (4) a binary correctness label (1 if the LLM prediction matches the gold label, 0 otherwise). These vectors are concatenated and passed to the proxy classifier, which is trained to predict correctness directly.

To collect the data, we ran standard MCTS (without the proxy) and sampled 30 unique prompts from different depths of the search tree to capture a range from early, simple prompts to more complex ones appearing later in the search. Each prompt was paired with 500 clauses from the training split of CLAUDETTE (see Table \ref{table:CLAUDETTE_stats}), with a balanced 50:50 distribution of \textit{fair} and \textit{unfair} clauses to ensure performance for the underrepresented \textit{unfair} class. For each (prompt, clause)-pair, we queried the LLM deterministically and assigned a correctness label based on the dataset's gold label. We also added the gold label as input to the correctness dataset, yielding 15,000 (prompt, clause, label)-triples for training. A validation set was built using the same procedure with 200 unseen clauses, sampled without enforcing label balance, intentionally sampled without enforcing label balance to preserve the natural distribution of LLM correctness and enable realistic evaluation.

During search inference, the trained proxy evaluates every (prompt, clause, label)-triple in the score set. If the proxy predicts \textit{correct}, we retain the gold label. If it predicts \textit{incorrect}, we flip it. The resulting sequence of predictions is compared to the gold labels, and the macro F1 score is used to estimate the performance of the prompt within the MCTS loop.


\subsubsection{Model Structure of the Prompt Scorer}


We tested two architectures for the proxy prompt evaluator: (1) a logistic regression classifier as used in \citet{polo2024efficientmultipromptevaluationllms}, and (2) a two-layer multilayer perceptron (MLP), inspired by \cite{Goodfellow-et-al-2016, afzal2025knowingsayingllmrepresentations}. For the logistic regression model we used the scikit-learn library implementation \cite{scikit-learn}. For the MLP classifier, we use a compact feed-forward neural network with three hidden layers of 512, 256, and 128 units, each using ReLU activation \cite{agarap2019deeplearningusingrectified} and dropout. The output layer is a single neuron with a sigmoid function for binary classification. More details of the model architectures can be found in Appendix \ref{app:PromptEval_Hyperparams}.

\subsubsection{Choice of Input Embeddings}


We experimented with two different embeddings to encode the input of the correctness dataset: 
\begin{itemize}[noitemsep]
    \item \textit{Sentence-BERT} (SB) \cite{reimers2019sentencebertsentenceembeddingsusing}, using the all-MiniLM-L6-v2 model \cite{reimers-2020-minilm} from the sentence-transformers library. For Sentence-BERT, text is tokenized and processed through the pre-trained model to generate 384-dimensional embeddings.

    
    \item \textit{Fine-tuned LEGAL-BERT} (FLB). We also experimented with more domain-specific and task-informed embedding. We fine-tune LEGAL-BERT \cite{chalkidis2020legalbertmuppetsstraightlaw} by training it on the fairness prediction task on the CLAUDETTE dataset. We take the [CLS] token representation from the final layer, resulting in 768-dimensional embeddings. More details of the model architectures can be found in Appendix \ref{app:FLB_Hyperparams}. 
    
\end{itemize}




We first conducted a preliminary study on embedding impact in Table \ref{table:embeddings}. We compared the performance of different embeddings when used with a logistic regression correctness classifier. Across both accuracy and macro F1, fine-tuned LEGAL-BERT embeddings yield the strongest results, with a validation accuracy of 0.93 and a macro F1 score of 0.93, outperforming Sentence-BERT. General-purpose embeddings like Sentence-BERT underperform against domain- and task-specific embeddings. The embedding choice therefore has a large impact on proxy model performance.

\begin{table}[t]
\centering
\begin{tabular}{ccc} 
 \hline
  & SB & FLB \\
 \hline
 Train accuracy & 0.94 & \textbf{0.94}\\
 Val accuracy & 0.85 & \textbf{0.93} \\ 
 \hline
 Train macro F1 & 0.94 & \textbf{0.94} \\
 Val macro F1 & 0.86 & \textbf{0.93} \\ 
 \hline
\end{tabular}
\caption{Logistic regression performance using different input embeddings}
\label{table:embeddings}
\end{table}

\begin{table}[h]
\centering
\begin{tabular}{ccc} 
 \hline
  & LogReg  & MLP  \\
 \hline
 SB & 0.85 & \textbf{0.93} \\
 FLB & \textbf{0.93} & 0.91 \\ 
 \hline
\end{tabular}
\caption{Validation accuracy of different classifier architectures.}
\label{table:LogReg_MLP}
\end{table}

While Table \ref{table:embeddings} shows that fine-tuned LEGAL-BERT is the strongest embedding for a linear proxy, Table \ref{table:LogReg_MLP} shows that the combination of embedding and architecture should also be considered. An MLP paired with Sentence-BERT matches the validation accuracy of 0.93 set by logistic regression with fine-tuned LEGAL-BERT. This result suggests that a non-linear scorer can extract more signal even from a general-purpose embedding, but gains less from the fine-tuned embeddings. 

In the following experiments, we tested our prompt optimization using the best two variants of the proxy scorer module (see Table \ref{table:LogReg_MLP}): (1) a logistic regression model with fine-tuned LEGAL-BERT embeddings, and (2) a MLP classifier with Sentence-BERT embeddings. 


\section{Evaluation Results}

In this section, we evaluate the effectiveness of our approach, which integrates MCTS with a proxy scorer, on the task of ToS fairness classification on the test set. We benchmark our method against three baselines categories: (1) traditional classifiers finetuned on the whole train set (SVM, BERT), (2) zero-shot LLM performance, and (3) baseline prompt optimization methods (OPRO, GRIPS). For all prompt optimization methods, we report the performance of the final highest-scoring prompts identified by each method. We report both accuracy and macro F1, as the test set is heavily class-imbalanced. In addition, we conduct an ablation study to assess the contribution of the proxy scorer. Finally, we complement the quantitative results with a qualitative analysis to offer a concrete insights of the improved prompts.

\begin{table}[t]
\centering
\resizebox{\columnwidth}{!}{%
\begin{tabular}{lcc} 
 \hline
  & Accuracy & Macro F1 \\
 \hline
 SVM w TD-IDF Vectorizer & 0.90 & 0.78 \\ 
 Fine-tuned LEGAL-BERT & 0.94 & 0.85 \\ 
 \hline
 Zero-Shot & 0.64 & 0.53 \\
 \hline
 GrIPS & 0.22 & 0.22 \\ 
 OPRO & 0.53 & 0.46 \\
 \textbf{MCTS w PromptEval-LogReg} & \textbf{0.90} & \textbf{0.69} \\
 \textbf{MCTS w PromptEval-MLP} & \textbf{0.90} & \textbf{0.73} \\
 \hline
\end{tabular}%
}
\caption{Binary fairness classification performance of prompt optimization approaches. SVM and BERT were trained on the whole training set.}
\label{table:overall_res}
\end{table}

\subsection{Overall Results}

Table \ref{table:overall_res} demonstrates our main results. Both versions of our approach outperformed the zero-shot, OPRO, and GrIPS baselines in binary classification, with the MLP-based variant reaching comparable performance to the SVM trained on the full dataset. Although the proxy-based methods did not surpass fine-tuned LEGAL-BERT models, they demonstrate that competitive performance can be achieved without large-scale training and with substantially lower computational cost. However, it is important to mention that the legal context provided to the LLM for scoring the refined prompts in our framework was richer than that used for the OPRO and GrIPS evaluations. This difference also influenced performance, as even the zero-shot baseline outperformed them.

\subsection{Ablation Experiments}


To ascertain the benefit of the proxy scorer, we isolate the contribution of the proxy scorer to overall optimization quality and efficiency by comparing MCTS variants that use full LLM-based scoring, with our PromptEval-based proxy variants.

As shown in Table \ref{table:overall_res} and \ref{table:proxy_abl}, MCTS with macro F1 achieves the highest scores, and its binary performance is comparable to the SVM trained on the full training split. It also outperforms OPRO and GrIPS, which lack error feedback, underscoring its importance. 

\begin{table}[h!]
\centering
\resizebox{\columnwidth}{!}{%
\begin{tabular}{lcc} 
 \hline
  & Accuracy & Macro F1 \\
 \hline
 MCTS w random scores & 0.81 & 0.67 \\
 MCTS w PromptEval-LogReg & 0.90 & 0.69 \\
 MCTS w accuracy scores & 0.87 & 0.72 \\
 MCTS w PromptEval-MLP & 0.90 & 0.73 \\
 MCTS w macro F1 scores & 0.89 & 0.74 \\
 \hline
\end{tabular}%
}
\caption{Binary fairness classification performance of MCTS with different scoring methods.}
\label{table:proxy_abl}
\end{table}

Although our MCTS implementation with the PromptEval-based scoring modules could not beat the best performing standard MCTS implementation, we still achieved an improvement over the random MCTS baseline. In particular, the prompt found by the MLP proxy model achieves comparable performance to the best prompt found through actual scoring. The reduction in sampling noise yields more stable average rewards than the limited score set in the standard MCTS approach. The execution time was also greatly reduced by a factor of 3. However, since we called the LLM via an API and ran the predictor model locally, it is hard to make universal claims about the speedup and efficiency.

\subsection{Qualitative Analysis}

\begin{figure}[h]
\centering
\includegraphics[width=0.99\columnwidth]{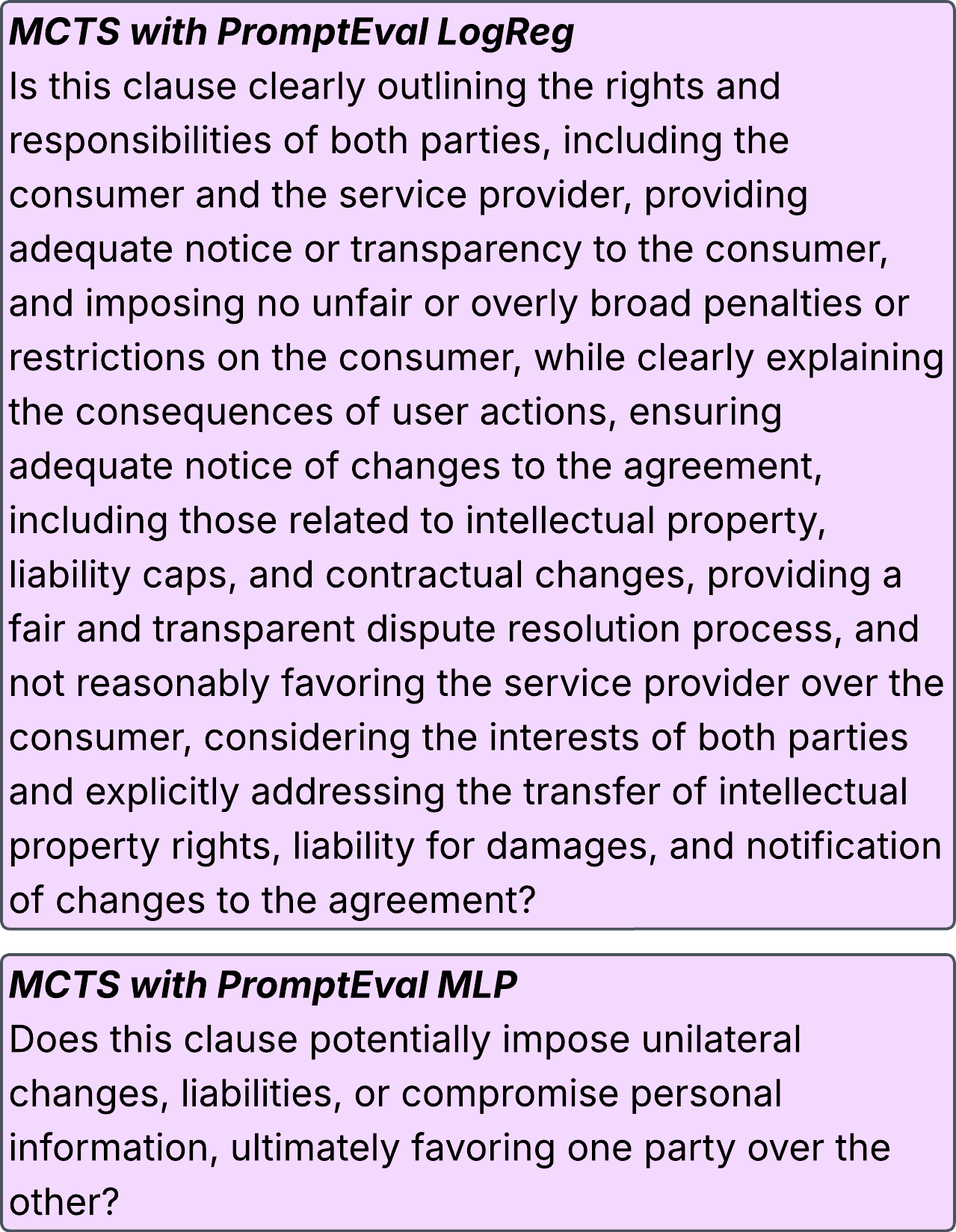}
\caption{Final prompts found with our approach.}
\label{fig:final_prompts}
\end{figure}

The initial prompt, as shown in \S \ref{init_prompt}, assumes the LLM inherently understands the legal concept of \textit{fairness} for the consumer in the context of ToS agreements without any explicit guidance. The prompts expanded by our approach (see Figure \ref{fig:final_prompts}) give the LLM more context on what exactly is meant by \textit{fairness}. It was also observed that the length of the final prompts depend on the tree depth that it was found at, since prompts tend to get longer with increasing depth.

Despite the stated advantages of using a proxy model during scoring, there are also limitations to consider. In particular, the computational burden was shifted from the MCTS runs to the construction of the dataset used to train our proxy model. If we quantify the cost in terms of expensive LLM calls and disregard other comparatively small factors like training, proxy model inference, and embedding generation with caching, we calculate the break-even point as follows.

The cost for creating the dataset is defined by 30 prompts combined with 500 samples, resulting in $30 \cdot 500 = 15,000$ LLM calls. The cost of a single expansion step in standard MCTS includes 20 calls for the gradient set, 2 calls to generate and apply the gradient, and 200 calls for evaluation on the score set. With 4 candidates per expansion, this totals $(20+2+200) \cdot 4 = 888$ calls. When using the proxy, the evaluation on the score set is replaced by proxy inference, reducing the total to $(20+2)\cdot 4 = 88$ calls (see Table \ref{table:llmcall_comparison}). 

\begin{table}[h!]
\centering
\resizebox{\columnwidth}{!}{%
\begin{tabular}{lcc} 
 \hline
 \multirow{2}{*}{Method (Score Set Size)} & Train Dataset & Expansion \\
  & Creation & Step \\
 \hline
 Standard MCTS (200) & - & 888 \\
 Standard MCTS (8,279) & - & 33,204  \\
 Proxy MCTS (8,279) & 15,000 & 88 \\
 \hline
\end{tabular}%
}
\caption{Comparison of LLM calls between standard and proxy MCTS.}
\label{table:llmcall_comparison}
\end{table}

To reach the break-even point, we therefore need to use at least $15,000 \div (888 - 88) = 18.75 < 19$ expansions. Our experiments show that the average number of expansions per MCTS run is 35 due to early stopping, indicating that the proxy approach becomes cost-efficient within a single run. The efficiency is further increased by the reusability of the proxy across MCTS runs and the ability to extend the score set without additional LLM calls.

Furthermore, the model is vulnerable to outliers, as it might fail to generalize, given the small number of prompts in the training data and the large potential search space of prompts. If the model vastly overestimates the performance of a certain prompt, this prompt is likely to be chosen as the final prompt, if we replace all scoring with our proxy. Since our chosen proxy variants exhibit black box characteristics, it becomes hard to detect biases during the fast scoring method.

\section{Conclusion}
\label{sec:conclusion}

In this paper, we propose augmenting a prompt optimization framework with a proxy prompt scorer. Our experimental results show that using a lightweight correctness prediction model as a proxy enables existing prompt optimization techniques to avoid repeatedly querying an LLM over the validation set, which is computationally expensive. In particular, the MLP-based proxy evaluator achieves performance close to the best-performing standard implementation that relies, while significantly reducing computation time and cost, which highlighting the effectiveness of our methodology. 

For future work, several promising directions can be explored. First, experimenting alternative proxy model architectures, including Transformers or Bayesian Models to better capture the interaction between prompt and task performance. Another promising direction is to optimize the score set using active or curriculum learning strategies. By selectively including the most informative or representative clauses, active learning can reduce the number of evaluations required while maintaining reliable performance estimates. Similarly, curriculum learning can improve the proxy model’s stability by starting with easier examples and progressively incorporating harder ones. These approaches would allow for more efficient and effective prompt evaluation, reducing computational cost while improving generalization to the full dataset.

\section{Limitations}

Our work was conducted on a small LLM with limited capabilities, making it heavily dependent on the legal context provided and sensitive to the precise wording of that context. The effectiveness of the proxy model is likewise tied to the LLM it was trained on and the error patterns specific to that model. Further research is needed to determine whether our findings generalize to larger models.

Additionally, due to budget constraints, our MCTS framework and PromptEval-based classifier relied only on a small subset of the training and validation data to train and to generate the search space. Using larger subsets may introduce more diversity and potentially improve performance.

Another limitation is the multi-label classification task of unfairness categories. Our quick scoring via proxy PromptEval-based models were only conducted on the binary classification task. Predicting correctness on multi-label classification is more difficult and it is left to see whether the proxy could sufficiently predict correctness to draw useful conclusions about the performance of a multi-label prompt.

\section*{Acknowledgments}
We thank the anonymous reviewers for valuable comments. SX is supported by the Independent Research Fund Denmark (DFF) ALIKE grant 4260-00028B.

\bibliography{custom}

\appendix
\section{Appendix}



\subsection{Hyperparameters of Proxy Prompt Evaluator }
\label{app:PromptEval_Hyperparams}
For our logistic regression classifier, we used the standard LBFGS solver that the Logistic Regression class from scikit-learn uses by default. We set the maximum number of optimization iterations to 1000 and the L2-regularization parameter C to 1.0, which is the default as well.

For our different MLP architectures, including the medium size one, we used an Adam optimizer with a learning rate of 0.001, dropout rate of 0.3 and batch size of 32. We trained with early stopping with a patience of 10 and a weight decay of 1e-4.

\subsection{Hyperparameters of finetuing LegalBERT }
\label{app:FLB_Hyperparams}
To get more domain-specific and task-informed embeddings, we finetune a LegalBERT \cite{chalkidis2020legalbertmuppetsstraightlaw} by training it on the fairness prediction task on the CLAUDETTE dataset. We trained the base architecture with a classification head on the task of binary fairness prediction for all training and validation clauses with cross-entropy loss. We used AdamW with a learning rate of 2e-5 and decay of 0.01. The model was trained for 3 epochs. To generate embeddings, we remove the classifier head and proceed the same way as with base LEGAL-BERT.


\end{document}